\documentclass[runningheads]{llncs}
\usepackage{graphicx}

\usepackage{tikz}
\usepackage{comment}
\usepackage{amsmath,amssymb} 
\usepackage{color}

\usepackage[breaklinks=true,colorlinks=false,bookmarks=false]{hyperref}
\DeclareUrlCommand\url{\color{magenta}}
\DeclareUrlCommand\nolinkurl{\color{magenta}}

\usepackage[accsupp]{axessibility}  
\usepackage{epsfig}
\usepackage{graphicx}
\usepackage{dsfont}
\usepackage{pifont}
\usepackage{enumitem}

\usepackage{xcolor}
\usepackage{amsfonts}
\usepackage{soul}
\usepackage{subfigure}
\usepackage{booktabs}
\usepackage{arydshln}
\usepackage{adjustbox}

\usepackage{tabularx}
\newcolumntype{L}[1]{>{\raggedright\arraybackslash}p{#1}}
\newcolumntype{C}[1]{>{\centering\arraybackslash}p{#1}}
\newcolumntype{R}[1]{>{\raggedleft\arraybackslash}p{#1}}

\begin{document}
\pagestyle{headings}
\mainmatter
\def\ECCVSubNumber{6599}  

\title{Learning Long-Term Spatial-Temporal Graphs for Active Speaker Detection} 

\makeatletter
\renewcommand*{\@fnsymbol}[1]{\ifcase#1\or$\ast$\else\@$\dagger$\fi}
\makeatother

\titlerunning{SPELL for Active Speaker Detection}
%
\author{Kyle Min$^{1}$\thanks{Authors contributed equally},
Sourya Roy$^{2\,\ast}$\thanks{Work partially done during an internship at Intel Labs},
Subarna Tripathi$^1$,
Tanaya Guha$^3$, and
Somdeb Majumdar$^1$}

\authorrunning{K. Min et al.}

\institute{
$^1$Intel Labs,\,
$^2$UC Riverside,\, 
$^3$University of Glasgow \\
\email{\{kyle.min,subarna.tripathi,somdeb.majumdar\}@intel.com}
}

\maketitle

\label{abs}
\begin{abstract}
Active speaker detection (ASD) in videos with multiple speakers is a challenging task as it requires learning effective audiovisual features and spatial-temporal correlations over long temporal windows. In this paper, we present SPELL, a novel spatial-temporal graph learning framework that can solve complex tasks such as ASD. To this end, each person in a video frame is first encoded in a unique node for that frame. Nodes corresponding to a single person across frames are connected to encode their temporal dynamics. Nodes within a frame are also connected to encode inter-person relationships. Thus, SPELL reduces ASD to a node classification task. Importantly, SPELL is able to reason over long temporal contexts for all nodes without relying on computationally expensive fully connected graph neural networks. Through extensive experiments on the AVA-ActiveSpeaker dataset, we demonstrate that learning graph-based representations can significantly improve the active speaker detection performance owing to its explicit spatial and temporal structure. SPELL outperforms all previous state-of-the-art approaches while requiring significantly lower memory and computational resources. Our code is publicly available: \url{https://github.com/SRA2/SPELL}

\end{abstract}

\section{Introduction}
\label{sec:intro}
Holistic scene understanding in the wild is still a challenge in computer vision despite recent breakthroughs in several other areas. A \emph{scene} represents real-life events spanning complex visual and auditory information, which are often intertwined.
Active speaker detection (ASD) is a key component in scene understanding and is an inherently multimodal (audio-visual) task. The objective here is, given a video input, to identify which persons are speaking in each frame. This has numerous practical applications ranging from speech enhancement systems~\cite{afouras2018conversation} to human-robot interaction~\cite{stefanov2016look,stefanov2017vision}.

\begin{figure}[t!]
\centering
\includegraphics[width=0.88\linewidth]{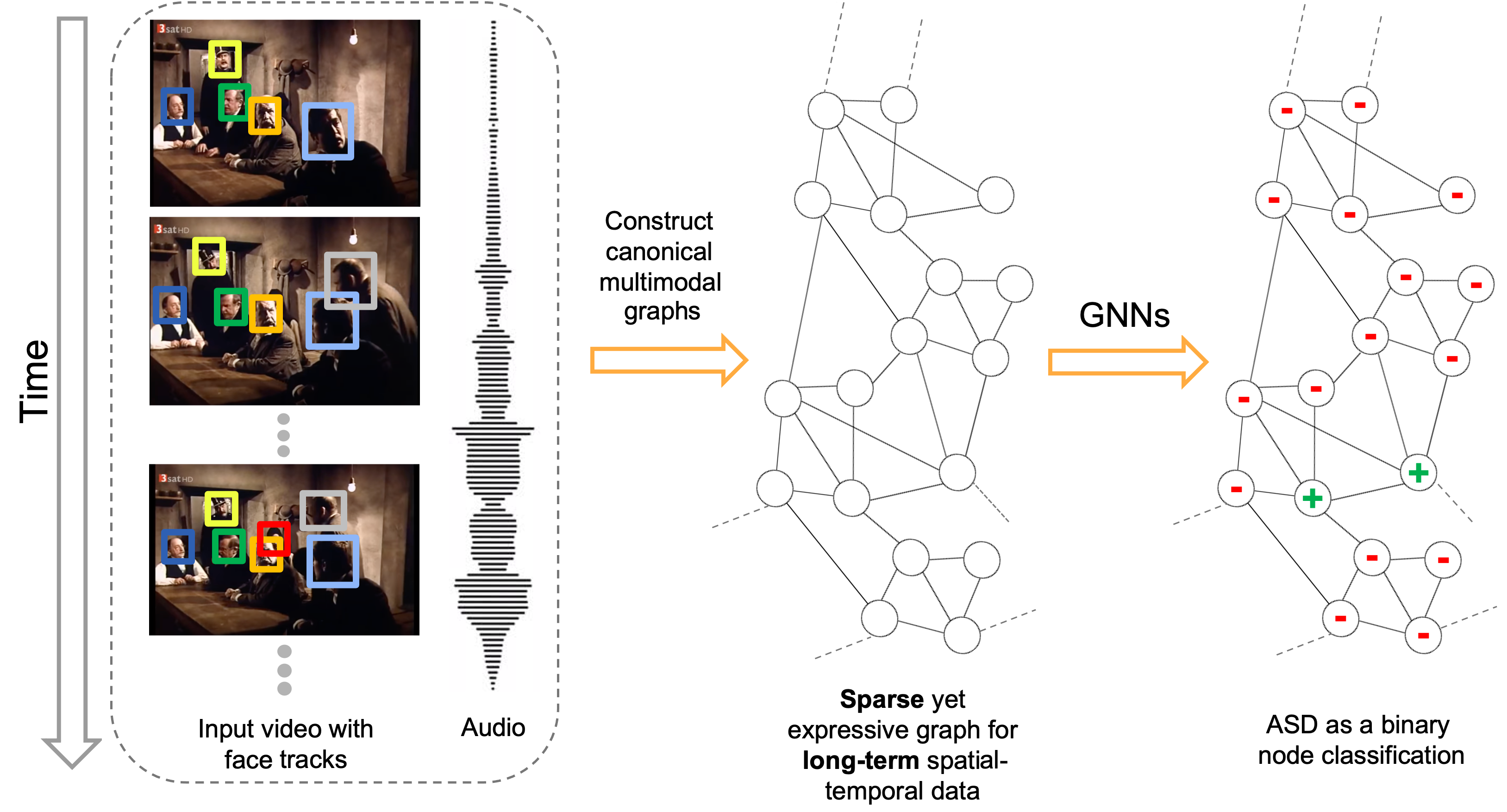}
  \caption{SPELL converts a video into a canonical graph from the audio-visual input data, where each node corresponds to a person in a frame, and an edge represents a spatial or temporal interaction between the nodes. The constructed graph is dense enough for modeling long-term dependencies through message passing across the temporally-distant but relevant nodes, yet sparse enough to be processed within low memory and computation budget. 
  The Active Speaker Detection (ASD) task is posed as a binary node classification in this long-range spatial-temporal graph.}
  \label{fig:teaser}
\end{figure}
Earlier efforts on ASD had limited success due to the unavailability of large datasets, powerful learning models, or computing resources \cite{cutler2000look,everingham2006hello,everingham2009taking}. With the release of AVA-ActiveSpeaker~\cite{roth2020ava}, a large and diverse ASD dataset, a number of promising approaches have been developed including both visual-only and audio-visual methods.
As visual-only methods \cite{everingham2006hello} are unable to distinguish between verbal and non-verbal lip movements, more recent approaches have focused on joint modeling of the audio-visual information. Audio-visual approaches \cite{alcazarActiveSpeakersContext2020,zhang2021unicon,tao2021someone,MAAS2021,ASDNet_ICCV2021} address the task by first encoding visual (primarily facial) and audio features from videos, and then by classifying the fused multimodal features. Such models generally have multi-stage frameworks \cite{alcazarActiveSpeakersContext2020,MAAS2021,zhang2021unicon,ASDNet_ICCV2021} and show good detection performance. However, state-of-the-art methods have relied on complex architectures
for processing the audio-visual features with high computation and memory overheads. For example, TalkNet~\cite{tao2021someone} suggests using a transformer-style architecture~\cite{vaswani2017attention} to model the cross-modal information from the audio-visual input. ASDNet~\cite{ASDNet_ICCV2021}, which is the leading state-of-the-art method, uses a complex 3D convolutional neural network (CNN) to extract more powerful features. These approaches are not scalable and may not be suitable for real-world situations with limited memory and computation budgets.

In this paper, we propose an efficient graph-based framework, which we call \href{https://github.com/kylemin/SPELL}{\color{magenta}\textbf{SPELL}} (\textbf{Sp}atial-T\textbf{e}mpora\textbf{l} Graph \textbf{L}earning). Figure~\ref{fig:teaser} illustrates an overview of our framework. We construct a multimodal graph from the audio-visual data and cast the active speaker detection as a graph node classification task. 
First, we create a graph where each node corresponds to each person at each frame and the edges represent spatial or temporal relationships among them. 
The initial node features are constructed using simple and lightweight 2D CNNs instead of a complex 3D CNN or a transformer. 
Next, we perform binary node classification -- active or inactive speaker -- on this graph by learning a three-layer graph neural network (GNN) model each with a small number of parameters. In our framework, graphs are constructed specifically for encoding the spatial and temporal dependencies among the different facial identities. Therefore, the GNN can leverage this graph structure and model the temporal continuity in speech as well as the long-term spatial-temporal context, while requiring low memory and computation.

Although the proposed graph structure can model the long-term spatial-temporal information from the audio-visual features, it is likely that some of the short-term information may be lost in the process of feature encoding. This is because we use 2D CNNs that are not well-suited for processing the spatial-temporal information when compared to the transformer or the 3D CNNs. To encode the short-term information, we adopt TSM~\cite{lin2019tsm} - a generic module for 2D CNNs that is capable of modeling temporal information without introducing any additional parameters or computation. We empirically verify that SPELL can benefit both from the supplementary short-term information provided by TSM and the long-term information modeled by our graph structure.

We show the effectiveness of SPELL by performing extensive experiments on the AVA-ActiveSpeaker dataset~\cite{roth2020ava}. Using our spatial-temporal graph framework on top of the TSM-inspired feature encoders, SPELL outperforms all previous state-of-the-art approaches. Critically, SPELL requires significantly less hardware resources for the visual feature encoding (0.7 GFLOPs, 11.2M \#Params) compared to ASDNet~\cite{ASDNet_ICCV2021} (13.2 GFLOPs, 48.6M params), which is the leading state-of-the-art method. In addition, SPELL achieved 2nd place in the AVA-ActiveSpeaker challenge at ActivityNet 2022\footnote{\url{https://research.google.com/ava/challenge.html}}, which also demonstrates the effectiveness of our method (please refer to the technical report~\cite{minintel}).

There are three main contributions in this paper:
\begin{itemize}[noitemsep,nolistsep]
    \item We present a graph-based approach for solving the task of active speaker detection 
    over long time supports  
    by casting it as a node classification problem. 
    \item Our model, SPELL, learns from videos to model the short-term and long-term spatial-temporal information. 
    Specifically, we propose to construct graphs on the TSM-inspired audio-visual features.
    The graphs are dense enough for message passing across temporally-distant nodes, yet sparse enough to model their interactions within tight memory and compute constraints.
    
    
    
   
   \item SPELL notably outperforms existing methods with lower memory and computation complexity on the active speaker detection benchmark dataset, AVA-ActiveSpeaker.

\end{itemize}








\section{Related Work}
\label{sec:related}

We discuss related works in two relevant areas: application of GNNs in video scene understanding and active speaker detection.

\textbf{GNNs for scene understanding. }
CNNs, Long Short Term Memory (LSTM), and their variants have long dominated the field of video understanding. In recent times, two new types of models are gaining traction in many areas of visual information processing: Transformers \cite{vaswani2017attention} and GNNs. They are not necessarily in competition with the former models, but it has been shown that they can augment the performance of CNN/LSTM based models. Applications of specialized GNN models in video understanding include visual relationship forecasting \cite{mi2021visual}, dialog modeling \cite{geng2020spatio}, video retrieval~\cite{tan2021logan}, emotion recognition \cite{shirian2020learnable}, and action detection \cite{zhang2019structured}. GNN-based generalized video representation frameworks have also been proposed \cite{arnab2021unified,nagarajan2020ego,patrick2021space} that can be used for multiple downstream tasks. For example, in Arnab \textit{et al.}~\cite{arnab2021unified}, a fully connected graph is constructed over the foreground nodes from video frames in a sliding window fashion, and a foreground node is connected to other context nodes from its neighboring frames. The message passing over the fully connected spatial-temporal graph is expensive in terms of the computational time and memory. Thus in practice such models end up using a small sliding window, making them unable to process longer-term  sequences. SPELL also operates on foreground nodes - particularly, faces. However, the graph structure is not fully connected. We construct the graph such that it enables interactions only between relevant nodes over space and time. The graph remains sparse enough such that the longer-term context can be accommodated within a comparatively smaller memory and compute budget.

\textbf{Active speaker detection (ASD). } Earlier work on active speaker detection by Cutler \textit{et al.} \cite{cutler2000look} detects correlated audio-visual signals using a time-delayed neural network. Subsequent works depend only on visual information and considers a simpler set-up focusing on lip and facial gestures \cite{everingham2006hello}. More recently, high-performing ASD models rely on large networks - developed for capturing the spatial-temporal variations in audio-visual signals, often relying on ensemble networks or complex 3D CNN features \cite{alcazarActiveSpeakersContext2020,tao2021someone}. Sharma \textit{et al.} \cite{sharma2020crossmodal} and Zhang \textit{et al.} \cite{zhang2019multi} both used large 3D CNN architectures for audio-visual learning. The Active Speaker in Context (ASC) model \cite{alcazarActiveSpeakersContext2020} uses non-local attention modules with an LSTM to model the temporal interactions between audio and visual features encoded by two-stream ResNet-18 networks~\cite{he2016deep}. TalkNet~\cite{tao2021someone} achieves superior performance through the use of a 3D CNN and a couple of Transformers~\cite{vaswani2017attention} resulting in an effectively large model. Another recent work, the ASDNet~\cite{ASDNet_ICCV2021}, uses 3D-ResNet101 for encoding visual data and SincNet~\cite{ravanelli2018speaker} for audio. The Unified Context Network (UniCon)~\cite{zhang2021unicon} proposes relational context modules to capture visual (spatial) and audio-visual context based on convolutional layers.
Much of these advances are due to the availability of the AVA-ActiveSpeaker dataset \cite{roth2020ava}. Previously available multimodal datasets (e.g. \cite{chakravarty2016cross}) were either smaller or constrained or lacked variability in data.  The work by Roth \textit{et al.} \cite{roth2020ava} also introduced a competitive baseline along with the large dataset. Their baseline involves jointly learning an audio-visual model that is end-to-end trainable. The audio and visual branches in this model are CNN-based which uses a depth-wise separable technique.

MAAS~\cite{MAAS2021} presents a different multimodal graph approach. Our work differs from MAAS in several ways, where the main difference is in the handling of temporal context. While MAAS focuses on short-term temporal windows to construct their graphs, we focus on constructing longer-term audio-visual graphs. More specifically, in MAAS, different faces are connected only between consecutive frames. In contrast, SPELL directly connects faces in a longer-term neighborhood controlled by the time threshold hyperparameter, $\tau$ (defined in Sec~\ref{sec:mmgraph}). In addition, SPELL exploits the temporal ordering patterns of the face tracks by using all the forward/backward/undirected edges in the time domain. 
In SPELL, each graph can span from 13 to 55 seconds (refer to Sec~\ref{long-term-context-reasoning}) of a video depending on the number of nodes. This is significantly larger than MAAS where the time window size is fixed at 1.59 seconds. During inference, SPELL performs single forward pass, whereas MAAS performs multiple forward passes.

\section{Method}
\label{sec:method}
\begin{figure}[t]
  \adjustbox{valign=t}{\begin{minipage}[t]{0.75\linewidth}
  \small
    \includegraphics[width=1\linewidth]{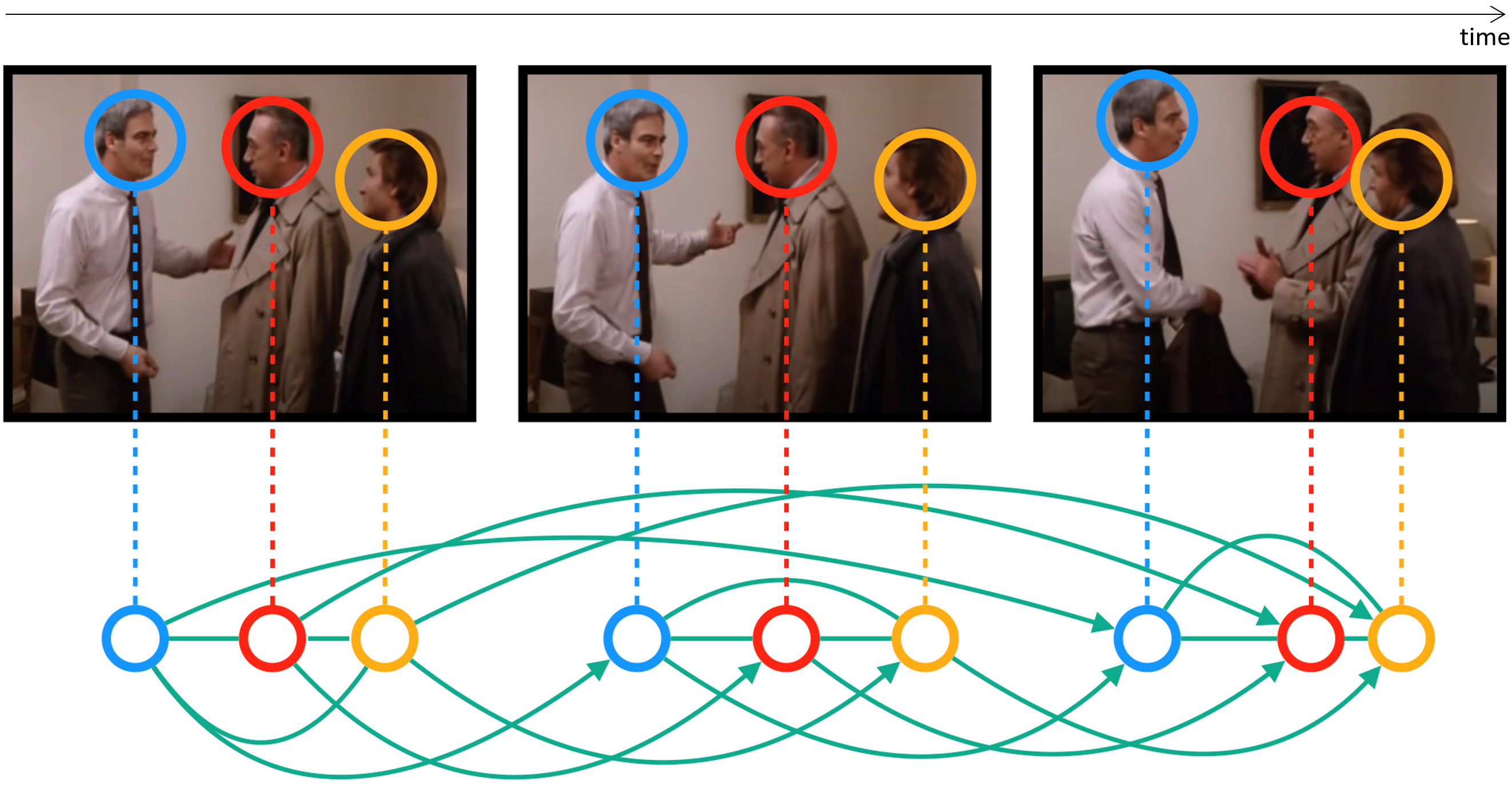}\\[-0.5ex] \hspace*{14.55em}(a)
  \end{minipage}}
  \vline \hspace{0em}
  \adjustbox{valign=t}{\begin{minipage}[t]{0.23\linewidth}
  \small
    \includegraphics[width=1\linewidth]{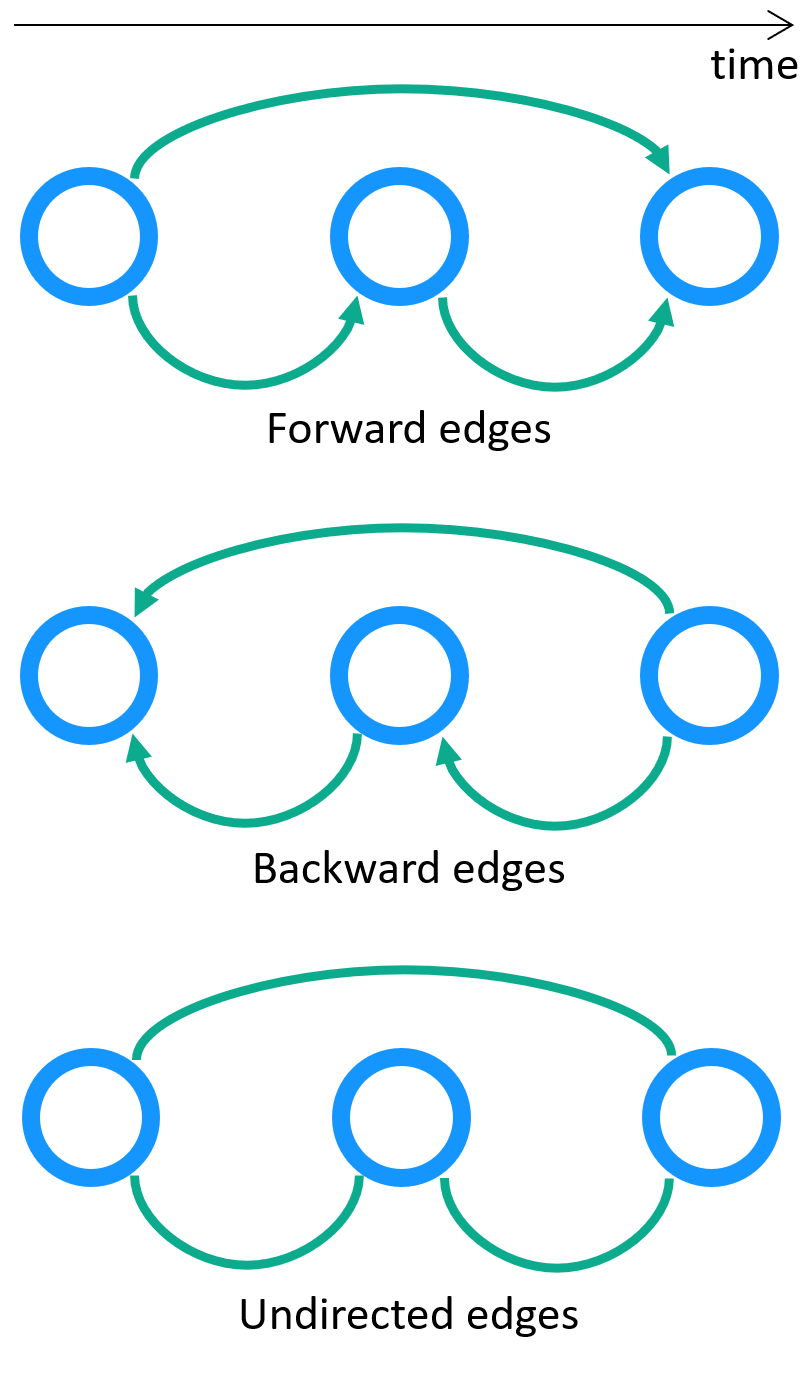}\\[-0.5ex] \hspace*{3.85em}(b)
  \end{minipage}}
  \caption{(a): An illustration of our graph construction process. The frames above are temporally ordered from left to right. The three colors of blue, red, and yellow denote three identities that are present in the frames. Each node in the graph corresponds to each face in the frames. SPELL connects all the inter-identity faces from the same frame with the undirected edges. SPELL also connects the same identities by the forward/backward/undirected edges across the frames. In this example, the same identities are connected across the frames by the forward edges, which are directed and only go in the temporally forward direction. (b): The process for creating the backward and undirected graph is identical, except in the former case the edges for the same identities go in the opposite direction and the latter has no directed edge. Each node also contains the audio information which is not shown here.}
  \label{fig:graph_construction}
\end{figure}

\noindent
In this section, we describe our approach in detail. Figure~\ref{fig:graph_construction} illustrates how SPELL constructs a graph from an input video where each node corresponds to a face within a temporal window of the video. SPELL is unique in terms of its canonical way of constructing the graph from a video. The graph is able to reason over long temporal contexts for all nodes without being fully-connected. This is an important design choice to reduce memory and computation overheads. The edges in the graph are only between \emph{relevant} nodes needed for message passing, leading to a sparse graph that can be accommodated within a small memory and computation budget. 
After converting the video into a graph, 
we train a lightweight GNN to perform binary node classification on this graph. The model architecture is illustrated in Figure~\ref{fig:model}. The model utilizes three separate GNN modules for the forward, backward, and undirected graph, respectively. Each module has three layers where the weight of the second layer is shared across all the above three modules. More details and the intuition behind the design choice are described in Section~\ref{subsec:arch}.


%

\subsection{Notations}
\noindent
Let $G=(V,E)$ be a graph with the node set $V$ and edge set $E$. For any $v\in V$, we define $N_v$ to be the set of neighbors of $v$ in $G$. We will assume the graph has self-loops, i.e., $v\in N_v$. In addition, let $X$ denote the set of given node features $\{\mathbf{x}_v \}_{v\in V}$ where $\mathbf{x}_v\in \mathbb{R}^d$ is the feature vector associated with the node $v$. Given this setup, we can define a $k$-layer GNN as a set of functions $\mathcal{F}=\{f_i\}_{i\in [k]}$ for $i\geq1$ where each $f_i:
V\rightarrow \mathbb{R}^m $ ( $m$ will depend on layer index $i$). All $f_i$  is parameterized by a set of learnable parameters. Furthermore, $X^{i}_V=\{\mathbf{x}_v\}_{v\in V}$ is the set of features at layer $i$ where $\mathbf{x}_v=f_i(v)$. Here, we assume that $f_i$ has access to the graph $G$ and the feature set from the last layer $X^{i-1}_V$. 
\begin{itemize}
    \item $\sf SAGE\text{-}CONV$ aggregation:
    This aggregation was proposed by \cite{hamilton2017inductive} and has a computationally efficient form. Given a $d$-dimensional feature set $X^{i-1}_V$
    , the function $f_i:V \rightarrow \mathbb{R}^m$ is defined for $i\geq1$ as follows:
    $$f(v)=\sigma \Big(\sum_{w\in  N_v}{\sf M}_i\mathbf{x}_w\Big)$$
 where $\mathbf{x}_w\in X^{i-1}_V$, ${\sf M}_i \in \mathbb{R}^{m\times d}$ is a learnable linear transformation, and $\sigma:\mathbb{R}\rightarrow\mathbb{R}$ is a non-linear activation function applied point-wise. 
    
    \item $\sf EDGE\text{-}CONV$ aggregation:
    $\sf EDGE\text{-}CONV$~\cite{wang2019dynamic} models global and local-structures by 
    applying channel-wise symmetric aggregation operation on the edge features associated with all the edges emanating from each node.
    The aggregation function $f_i:V\rightarrow \mathbb{R}^{m}$ can be defined as:
    $$f_i(v)=\sigma \Big(\sum_{w\in  N_v}{\sf g}_i\big( \mathbf{x}_v\circ \mathbf{x}_w
 \big)	\Big)$$ where $\circ$ denotes concatenation and ${\sf g}_i: \mathbb{R}^{2d}\rightarrow \mathbb{R}^m $ is a 
learnable transformation. Often ${\sf g}_i$ is implemented by MLPs. 
The number of parameters for $\sf EDGE\text{-}CONV$ is larger than $\sf SAGE\text{-}CONV$. This gives the $\sf EDGE\text{-}CONV$ layer more expressive power at a cost of 
higher complexity
and possible risk of overfitting. For our model, we set ${\sf g}_i$ to be an MLP with two layers of linear transformation and a non-linearity. We describe the details in section~\ref{sec:results}.  
    
\end{itemize}

\subsection{Video as a multimodal graph} \label{sec:mmgraph}
\noindent
We represent a video as a multimodal graph that is suitable for the task of active speaker detection.
We assume that the bounding-box information of every face region in each frame is given as per the problem set up. 
For simplicity, we assume that the entire video is represented by a single graph - if the video has $n$ faces in it, the graph will have $n$ nodes. 
In our actual implementation, we temporally order the set of all faces in a video, divide them in contiguous sets, and then construct one graph for each such set.

Let $B$ be the set of all face images cropped from an input video (i.e. face-crops). Then, each element $b\in B$ can be represented by a tuple $\sf(Box,Time,Id)$, where $\sf Box$ is the normalized bounding-box coordinates of a face-crop in its frame, $\sf Time$ is the time-stamp of its frame, and $\sf Id $ is a unique string that is common to all the face-crops that shares the same identity.

In other words, $B$ can be represented by a set of nodes $[n]$ where $n=|B|$ is the total number of faces that appear in the video.
${\sf Box}$ is treated as a map such that ${\sf Box}(i)$ is defined by the bounding-box coordinates of the $i$-th face for any $i\in [n]$. Similarly, ${\sf{Time}}(i)$ and ${\sf{Id}}(i)$ correspond to the time and identity of the $i$-th face, respectively. With this setup, the node set of $G=(V,E)$ is $V=[n]\cong B$, and for any $(i,j)\in [n]\times [n]$, we have $(i,j)\in E$ if either of the following two conditions are satisfied:

\begin{itemize}[itemindent=2.65cm]
    \setlength\itemsep{0.3em}
    \item ${\sf Id}(i)={\sf Id}(j)$ and $|{\sf Time}(i)$-${\sf Time}(j)| \leq \tau$
    \item ${\sf Time}(i)={\sf Time}(j)$
\end{itemize}
where $\tau$ is a hyperparameter for the maximum time difference between the nodes having the same identities. In essence, we connect two nodes (faces) if they share the same identity and are temporally close or if they belong to the same frame. Thus, the interactions between different speakers and the temporal variations of the same speaker can jointly be modeled. 

To pose the active speaker detection task as a node classification problem, we also need to specify the feature vectors for each node $v\in V$. We use a two-stream 2D ResNet~\cite{he2016deep} architecture as in \cite{roth2020ava,alcazarActiveSpeakersContext2020} for extracting the visual features of each face-crop and the audio features of each frame. Then, a feature vector of node $v$ is defined to be $x_v=[v_{{\sf visual}}\circ v_{{\sf audio}}]$ where $v_{{\sf visual}}$ is the visual feature of face-crop $v$ and $v_{{\sf audio}}$ is the audio feature of $v$'s frame where $\circ$ denotes the concatenation. Finally, we can write $G=(V,E,X)$ where $X$ is the set of the node features.

\begin{figure*}[t]
 \center
  \includegraphics[width=\textwidth]{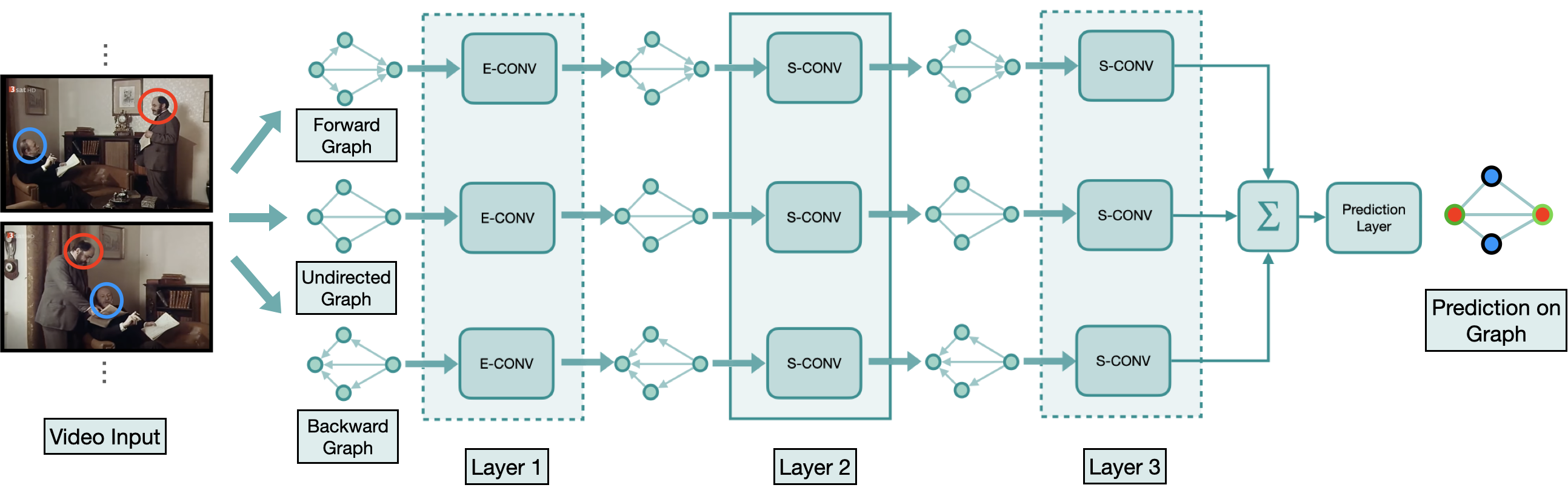}
  \caption{An illustration of our proposed \emph{Bi-directional} (a.k.a. \emph{Bi-dir}) GNN model for active speaker detection. Here, we have three separate GNN modules for the forward, backward, and undirected graph, respectively. Each module has three layers where the weight of the second layer is shared by all three graph modules. 
  The second layer is placed inside a solid-lined box to indicate the weight sharing while for the first and the third layer we use dotted-lines. E-CONV and S-CONV are shorthand for ${\sf EDGE}$-$\sf CONV$ and $\sf SAGE$-$\sf CONV$, respectively.
  We use the color coding: blue and red to denote different identities in input frames. The output of the third layers are added together and then passed to the prediction layer. It applies the sigmoid function to the summed features of every node and produces node classification probabilities.}
  \label{fig:model}
\end{figure*}

\subsection{ASD as a node classification task} \label{subsec:task}
\noindent
In the previous section, we described our graph construction procedure that converts a video into a graph $G=(V,E,X)$ where each node has its own audio-visual feature vector. During the training process, we have access to the ground-truth labels of all face-crops indicating if each of the face-crop is active speaker or not. Therefore, the task of active speaker detection can be naturally posed as a binary node classification problem in the constructed graph $G$, whether a node is speaking or not speaking. Specifically, we train a three-layer GNN for this classification task. The first layer in the network uses ${\sf EDGE\text{-}CONV}$ aggregation to learn pair-wise interactions between the nodes. For the last two layers, we observe that using ${\sf SAGE\text{-}CONV}$ aggregation provides better performance than ${\sf EDGE\text{-}CONV}$, possibly due to ${\sf EDGE\text{-}CONV}$'s tendency to overfit.

\subsection{SPELL} \label{subsec:arch}
We now describe how our graph construction and embedding strategy takes temporal ordering into consideration. Specifically, as we use the criterion: $|{\sf Time}(i)-{\sf Time}(j)| \leq \tau$ for connecting the nodes having the same identities across the frames, the resultant graph becomes undirected. In this process, we lose the information of the temporal ordering of the nodes. To address this issue, we explicitly incorporate temporal direction as shown in Figure~\ref{fig:graph_construction}(b). The undirected GNN is augmented with two other parallel networks; one for going forward in time and another for going backward in time. 

More precisely, in addition to the undirected graph, we create a forward graph where we connect $(i,j)$ if and only if $0\geq {\sf Time}(i)-{\sf Time}(j) \geq- \tau$. Similarly, $(i,j)$ is connected in a backward graph if and only if $0\leq {\sf Time}(i)-{\sf Time}(j) \leq \tau$. This gives us three separate graphs where each of the graphs can model different spatial-temporal relationships between the nodes. Furthermore, the weights of the second layer of each graph is shared across the three graphs. This weight sharing technique can enforce the temporal consistencies among the different information modeled by the three graphs as well as reduce the number of parameters. For the remaining parts of this paper, we will refer to this network that is augmented with the foward/backward graphs as \emph{Bi-directional} or \emph{Bi-dir} for short. 

The proposed three-layer \emph{Bi-dir} is illustrated in Figure~\ref{fig:model}. We note that right before the \emph{Bi-dir} is applied, the audio and the visual features are further encoded by two learnable MLP layers (linear transformation with ReLU activation) separately and then added to form the fused features for the graph nodes. After the fused features are processed by the first and the second layers, the third layer aggregates all the information and reduce the feature dimension to 1. These 1D features coming from the three separate graphs are added and applied to the sigmoid function to get the final prediction score for each node.

\subsection{Feature learning}\label{subsec:aug}
Similar to ASC~\cite{alcazarActiveSpeakersContext2020}, we use a two-stream 2D ResNet~\cite{he2016deep} architecture for the audio-visual feature encoding. The networks take as visual input $k$ consecutive face-crops and take as audio input the Mel-spectrogram of the audio wave sliced along the time duration of the face-crops for the visual stream. Although the 2D ResNet requires significantly lower hardware resources than 3D CNN counterparts or a transformer-style architecture~\cite{vaswani2017attention}, it is not specifically designed for processing spatial-temporal information that is crucial in understanding video contents. To better encode the spatial-temporal information, we augment the visual feature encoder with TSM~\cite{lin2019tsm}, which provides 2D CNNs with a capability to model the short-term temporal information without introducing any additional parameters or computation. This additional use of TSM can greatly improve the quality of the visual features, and we empirically establish that SPELL benefits from the supplementary short-term information. The audio-visual features from the two stream are concatenated to be node features $\{\mathbf{x}_v \}$.

\textbf{Data augmentation. } Reliable ASD models should be able to detect speaking signals even if there is a noise in the audio. To make our method robust to noise, we make use of data augmentation methods while training the feature extractor. 
Inspired by TalkNet~\cite{tao2021someone}, we augment the audio data by negative sampling. For each audio signal in a batch, we randomly select another audio sample from the whole training dataset and add it after decreasing its volume by a random factor. This technique can effectively increase the amount of training samples for the feature extractor by selecting negative samples from the whole training dataset.

\textbf{Spatial feature. } The visual features encoded by the 2D ResNet do not have any information about where each face is spatially located in each frame because we only use the cropped face regions in the visual feature encoding. Here, we argue that the spatial locations of speakers can be another type of inductive bias. In order to exploit the spatial information of each face-crop, we incorporate the spatial features corresponding to each face as additional input to the node feature as follows: We project the $4$-D spatial feature of each face region parameterized by the normalized center location, height and width ($x$, $y$, $h$, $w$) to a 64-D feature vector using a single fully-connected layer. The resulting spatial feature vector is then concatenated to the visual feature at each node.



\section{Experiments}
\label{sec:results}
We perform experiments on the large-scale AVA-ActiveSpeaker dataset~\cite{roth2020ava}. Derived from Hollywood movies, this dataset comes with a number of face tracks for active and inactive speakers and their audio signals. Its extensive annotations of the face tracks is a key feature that was missing in its predecessors.

\textbf{Implementation details. }
Following ASC~\cite{roth2020ava}, we utilize a two-stream network with a ResNet~\cite{he2016deep} backbone for the audio-visual feature encoder. In the training process, we perform visual augmentation including horizontal flipping, color jittering, and scaling and audio augmentation as described in Section~\ref{subsec:aug}. We extract the encoded audio, visual, and spatial features for each face-crop to make the node feature. For the visual features, we use a stack of 11 consecutive face-crops (resolution: 144$\times$144). We implement SPELL using PyTorch Geometric library~\cite{fey2019fast}. Our model consists of three GCN layers, each with 64 dimensional filters. The first layer is implemented by an $\sf EDGE$-$\sf CONV$ layer that uses a two-layer MLP for feature projection. The second and third GCN layers are of type $\sf SAGE$-$\sf CONV$ and each of them uses a single MLP layer. We set the number of nodes $n$ to 2000 and $\tau$ parameter to 0.9, which ensures that each graph fully spans each of the face tracks. We train SPELL with a batch size of 16 using the Adam optimizer~\cite{kingma2014adam}. The learning rate starts at $5 \times 10^{-3}$ and decays following the cosine annealing schedule~\cite{loshchilov2016sgdr}. The whole training process of 70 epochs takes less than an hour using a single GPU (TITAN V).


\begin{table}[t]
\centering
\caption{Performance comparisons with other state-of-the-art methods on the validation set of AVA-ActiveSpeaker datset~\cite{roth2020ava}. We report mAP (mean average precision). SPELL outperforms all the previous approaches. 3D Conv denotes an additional use of one or more 3D convolutional layers. Note that TSM~\cite{lin2019tsm} does not increase memory usage nor the computation cost.}
\begin{tabular}{L{2.8cm}|L{6.0cm}|C{1.8cm}}
\toprule
\, \textbf{Method} & \, \textbf{Feature encoding network} & \textbf{mAP(\%)} \\ \midrule
\, Roth \textit{et al.}~\cite{roth2020ava} & \, MobileNet~\cite{howard2017mobilenets} & 79.2 \\
\, Zhang \textit{et al.}~\cite{zhang2019multi} & \, 3D ResNet-18~\cite{stafylakis2017combining} + VGG-M~\cite{chatfield2014return} & 84.0 \\
\, MAAS-LAN~\cite{MAAS2021} & \, 2D ResNet-18~\cite{he2016deep} & 85.1 \\
\, Chung \textit{et al.}~\cite{chung2019naver} & \, VGG-M~\cite{chatfield2014return} + 3D Conv & 85.5 \\
\, ASC~\cite{alcazarActiveSpeakersContext2020} & \, 2D ResNet-18~\cite{he2016deep} & 87.1 \\
\, MAAS-TAN~\cite{MAAS2021} & \, 2D ResNet-18~\cite{he2016deep} & 88.8 \\
\, UniCon~\cite{zhang2021unicon} & \, 2D ResNet-18~\cite{he2016deep} & 92.0 \\
\, TalkNet~\cite{tao2021someone} & \, 2D ResNet-18/34~\cite{he2016deep} + 3D Conv & 92.3 \\
\, ASDNet~\cite{ASDNet_ICCV2021} & \, 3D ResNeXt-18~\cite{kopuklu2019resource} + SincDSNet~\cite{ravanelli2018speaker} & 93.5 \\ \midrule
\, SPELL (Ours) & \, 2D ResNet-18-TSM~\cite{he2016deep,lin2019tsm} & \textbf{94.2} \\
\, SPELL\texttt{+} (Ours) & \, 2D ResNet-50-TSM~\cite{he2016deep,lin2019tsm} & \textbf{94.9} \\
\bottomrule
\end{tabular}
\label{tab:main_table}
\end{table}

\subsection{Comparison with the state-of-the-art}

We summarize the performance comparisons of SPELL with other state-of-the-art approaches on the validation set of the AVA-ActiveSpeaker dataset~\cite{roth2020ava} in Table~\ref{tab:main_table}. We want to point out that SPELL significantly outperforms all the previous approaches using the two-stream 2D ResNet-18~\cite{he2016deep}. Critically, SPELL's visual feature encoding has significantly lower computational and memory overhead ($0.7$ GFLOPs and 11.2M parameters) compared to 
ASDNet~\cite{ASDNet_ICCV2021} (13.2 GFLOPs, 48.6M \#Params), the leading state-of-the-art method. A concurrent and closely related work MAAS~\cite{MAAS2021} also uses a GNN-based framework. MAAS-LAN uses a graph that is generated on a short video clip. To improve the detection performance, MAAS-TAN extends MAAS-LAN by connecting the graphs over time, which makes 13 temporally-linked graph spanning about 1.59 seconds. This time span is relatively shorter than SPELL since the SPELL graph spans around 
$13$-$55$ seconds, as explained in the next subsection. In addition, SPELL requires a single forward pass when MAAS performs multiple forward passes for each inference process.

\begin{table}[t]
\centering
\caption{Performance comparison of context-reasoning with state-of-the-art methods. SPELL without TSM~\cite{lin2019tsm} demonstrates the higher context-reasoning capacity of our method when compared to the other 2D CNN-based approaches.}
\begin{tabular}{L{4.2cm}|C{2.4cm}|C{2.4cm}|C{2.4cm}}
\toprule
\, \textbf{Method} & \textbf{Stage-1 mAP} & \textbf{Final mAP} & \textbf{$\Delta$mAP} \\ \midrule
\, MAAS-LAN~\cite{MAAS2021} & 79.5 & 85.1 & 5.6\\
\, ASC~\cite{alcazarActiveSpeakersContext2020} & 79.5 & 87.1 & 7.6\\ 
\, MAAS-TAN~\cite{MAAS2021} & 80.2 & 88.8 & 8.6\\
\, Unicon~\cite{zhang2021unicon} & 84.0 & 92.0 & 8.0 \\
\, ASDNet~\cite{ASDNet_ICCV2021} & \textbf{88.9} & 93.5 & 4.6\\
\, SPELL (Ours) & 88.0 & \textbf{94.2} & 6.2 \\
\, SPELL (Ours) w/o TSM & 82.6 & 92.0 & \textbf{9.4} \\
\bottomrule
\end{tabular}
\label{tab:context_capacity}
\end{table}

\subsection{Context-reasoning capacity} \label{long-term-context-reasoning}
Most of the previous approaches have multi-stage frameworks, which includes a feature-encoding stage for audio-visual feature extraction that is followed by one or more context-reasoning stages for modeling long-term interactions and the context information. For example, SPELL has a single context-reasoning stage that uses a three-layer \emph{Bi-dir} GNN for modeling long-term spatial-temporal information. In Table~\ref{tab:context_capacity}, we compare the performance of context-reasoning stages with previous methods. Specifically, we analyze the detection performance when using only the feature-encoding stage (Stage-1 mAP) and the final performance. The difference between the two scores can provide a good insight on the capacity of the context-reasoning modules. Because ASDNet~\cite{ASDNet_ICCV2021} uses 3D CNNs, it is likely that some degree of the temporal context is already incorporated in the feature-encoding stage, which leads to a low context-reasoning performance. Similarly, using TSM~\cite{lin2019tsm} provides the short-term context information in the feature-encoding stage,
which leads to a smaller score difference between the Stage-1 and Final mAP and thus underestimates the context-reasoning capacity.
Therefore, we also estimate the performance of SPELL without TSM. 
In this case, the context-reasoning performance of SPELL outperforms all the other methods, which shows the higher context-reasoning capacity of our method, thanks to the longer-term context modeling. Note that although ASC~\cite{alcazarActiveSpeakersContext2020}, MAAS~\cite{MAAS2021}, Unicon~\cite{zhang2021unicon}, and SPELL use the same 2D ResNet-18~\cite{he2016deep}, their Stage-1 mAP can be different due to the inconsistency of input resolution, number of face-crops, and training scheme.

\par
\textbf{Long-term temporal context. }
Note that $\tau$ (= 0.9 second in our experiments) in SPELL imposes \emph{additional} constraint on direct connectivity across temporally distant nodes. The face identities across consecutive time-stamps are always connected. 
Below is the estimate of the effective temporal context size of SPELL. 
AVA-ActiveSpeaker dataset contains $3.65$ million frames and $5.3$ million annotated faces, resulting into $1.45$ faces per frame. 
With an average of $1.45$ faces per frame, a graph with 500 to 2000 faces in sorted temporal order spans over $345$ to $1379$ frames which correspond to 13 to 55 seconds for a 25-fps video. 
In other words, the nodes in the graph might have a time-difference of about 1 minute, and SPELL is able to reason over that long-term temporal window within a limited memory and compute budget, thanks to the effectiveness of the proposed graph structure.  
It is note worthy that the temporal window size in MAAS~\cite{MAAS2021} is $1.9$ seconds and TalkNet~\cite{tao2021someone} uses up to $4$ seconds as long-term sequence-level temporal context. 

\begin{table}[t]
\centering
\caption{Complexity comparisons of the context-reasoning stage. SPELL achieves the best performance while requiring the lowest memory and computation consumption.}
\label{tab:modelsize}
\begin{tabular}{L{2.9cm}|C{2.4cm}|C{2.4cm}|C{2.4cm}}
\toprule
\, \textbf{Method}  & \textbf{\#Params(M)} & \textbf{Size(MB)} & \textbf{mAP(\%)}\\ 
\midrule
\, ASC~\cite{alcazarActiveSpeakersContext2020} & 1.13 & 4.32 & 87.1  \\
\, MAAS-TAN~\cite{MAAS2021} & 0.16 & 0.63 & 88.8 \\
\, ASDNet~\cite{ASDNet_ICCV2021} & 2.56 & 9.77 & 93.5 \\
\, SPELL (Ours) & 0.11 & 0.45 & 94.2 \\ 
\bottomrule
\end{tabular}
\end{table}

\subsection{Efficiency of the context-reasoning stage}
In Table~\ref{tab:modelsize}, we compare the complexity of the context-reasoning stage of SPELL with ASC~\cite{alcazarActiveSpeakersContext2020}, MAAS-TAN~\cite{MAAS2021}, and ASDNet~\cite{ASDNet_ICCV2021}. These methods release the source code for their models, so we use the official code to compute the number of parameters and the model size of the context-reasoning stage. ASC has about 10 times more parameters and model size than ours. Nevertheless, SPELL achieves $7.1\%$ higher mAP than ASC. SPELL has fewer number of parameters than MAAS-TAN even while achieving $5.4\%$ higher mAP. When compared to the leading state-of-the-art method, ASDNet, SPELL is one order of magnitude more computationally efficient.

\subsection{Ablation study}
We perform an ablative study to validate the contributions of individual components, namely TSM, Cx-reason (context-reasoning only with an undirected graph), \emph{Bi-dir} (augmenting context-reasoning with the forward/backward graphs), Audio-aug (audio data augmentation), and Sp-feat (spatial features). We summarize the main contributions in Table~\ref{tab:ablation}. We can observe that TSM, \emph{Bi-dir} graph structure, and audio data augmentation play significant roles in boosting the detection performance. This implies that 1) retaining short-term information in the feature-encoding stage is important, 2) processing the spatial-temporal information using our graph structure is effective, and 3) the negatively sampled audio makes our model more robust to the noise. Additionally, the spatial features also bring meaningful performance gain.

\begin{table}[t]
\centering
\caption{Performance comparisons of different ablative settings: TSM~\cite{lin2019tsm}, Cx-reason (context-reasoning only with an undirected graph), \emph{Bi-dir} (augmenting with forward/backward graphs), Audio-aug (audio data augmentation), Sp-feat (spatial features).}
\label{tab:ablation}
\begin{tabular}{C{1.8cm}C{1.8cm}C{1.8cm}C{1.8cm}C{1.8cm}C{2.3cm}}
\toprule
\textbf{TSM} & \textbf{Cx-reason}  & \textbf{\emph{Bi-dir}} & \textbf{Audio-aug} & \textbf{Sp-feat} & \textbf{mAP(\%)} \\ 
\midrule
- & - & - & - & - & 80.2 \\
- & - & - & \checkmark & - & 82.6 \\
\checkmark & - & - & \checkmark & - & 88.0 \\
\checkmark & \checkmark & - & \checkmark & - & 92.4 \\
\checkmark & \checkmark & \checkmark & \checkmark & - & 93.9 \\
\checkmark & \checkmark & \checkmark & \checkmark & \checkmark & 94.2 \\
\bottomrule
\end{tabular}
\end{table}

\begin{figure}[t]
  \adjustbox{valign=t}{\begin{minipage}[t]{0.47\columnwidth}
  \small
    \includegraphics[width=0.98\columnwidth]{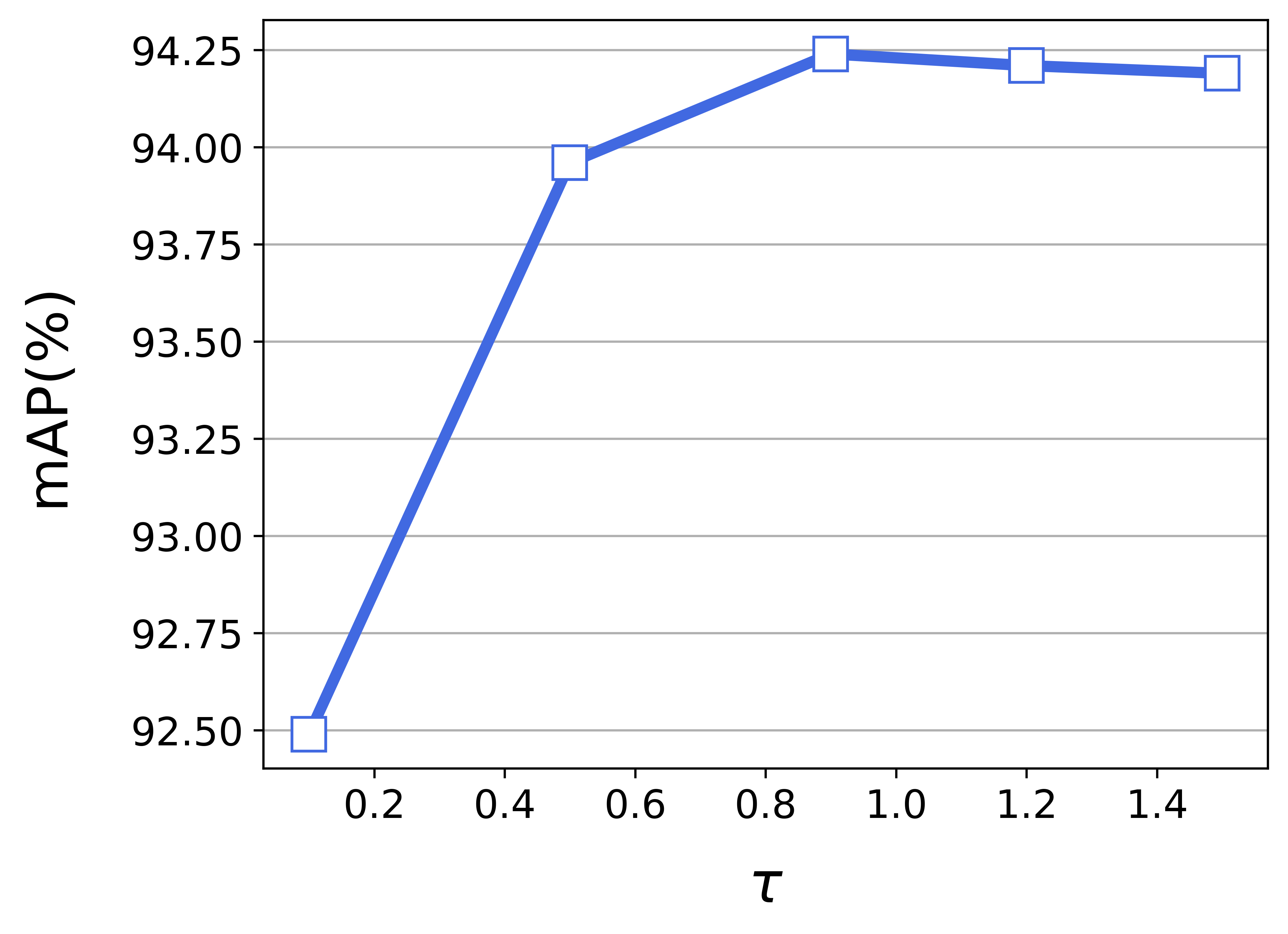}\\[-0.5ex] \hspace*{10.05em}(a)
  \end{minipage}}
  \hspace{0.02cm}
  \adjustbox{valign=t}{\begin{minipage}[t]{0.47\columnwidth}
  \small
    \includegraphics[width=0.98\columnwidth]{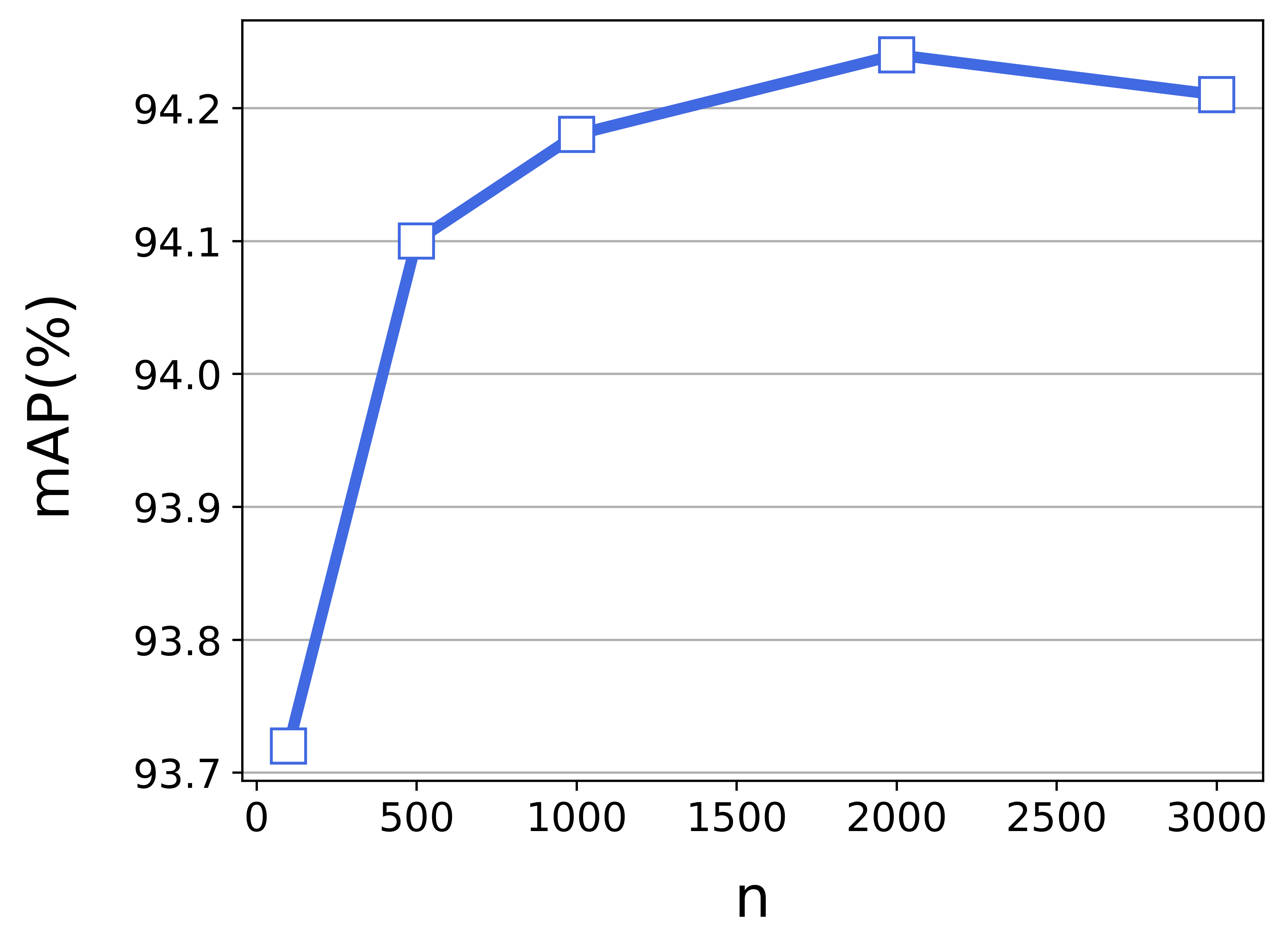}\\[-0.5ex] \hspace*{9.85em}(b)
  \end{minipage}}
  \caption{Study on the impact of two hyperparameters, which are $\tau$ (when $n$ is set to 2000) and $n$ (when $\tau$ is fixed at 0.9).}
  \label{fig:4}
\end{figure}


In addition, we analyze the impact of two hyperparameters: $\tau$ (Section~\ref{sec:mmgraph}) and the number of nodes in a graph embedding process. $\tau$ controls the connectivity or the edge density in the graph construction. Specifically, larger values for $\tau$ allow us to model longer temporal correlations but increases the average degree of the nodes, thus making the system more computationally expensive. In Figure~\ref{fig:4}, we can observe that connecting too distant face-crops deteriorates the detection performance. One potential reason behind this could be that the aggregation procedure becomes too smooth due to the high degree of connectivity. Interestingly, we also found that a larger number of nodes does not always lead to higher performance. This might be because after a certain point, larger number of nodes leads to over-fitting.




\begin{table}[t]
\centering
\caption{Comparisons of the detection performance and the model size with different filter dimensions.}
\begin{tabular}{C{2.4cm}|C{2.4cm}|C{2.4cm}|C{2.4cm}}
\toprule
\textbf{Filter Dim} & \textbf{\#Params(M)} & \textbf{Size(MB)} & \textbf{mAP(\%)} \\
\midrule
16 & 0.02 & 0.10 & 93.5 \\
32 & 0.05 & 0.21 & 93.9 \\
64 & 0.11 & 0.45 & 94.2 \\
128 & 0.29 & 1.14 & 94.1 \\
256 & 0.88 & 3.38 & 94.1 \\
\bottomrule
\end{tabular}
\label{tab:fdim}
\end{table}

We perform additional experiments with different filter dimensions of the ${\sf EDGE}$-$\sf CONV$ and $\sf SAGE$-$\sf CONV$. In Table~\ref{tab:fdim}, we show how the detection performance and the model size change depending on the filter dimension. We can observe that increasing the filter dimension above 64 does not bring any performance gain when the model size increases significantly.

We also perform an ablation study of the input modalities. When using only the visual features, the detection performance drops significantly from 94.2\% to 84.9\% mAP (when using only the audio: 55.6\%), which shows that both the audio and video modalities are important for this application.

\subsection{Qualitative analysis}
In the supplementary material, we show several detection examples to provide a qualitative analysis. The selected frames have multiple faces and have a long time-span about 5-10 seconds. In all of the provided examples in the supplementary material, SPELL correctly classifies all the speakers when the counterpart fails to do. The qualitative analysis demonstrates that SPELL is effective and that it is good at modeling spatial-temporal long-term information.

\section{Conclusion}
\label{sec:Conclusions}

We proposed SPELL - an effective graph-based approach to active speaker detection in videos. The main idea is to capture the long-term spatial and temporal relationships among the cropped faces through a graph structure that is aware of the temporal order of the faces. SPELL outperforms all the previous approaches and requires significantly less hardware resources when compared to the leading state-of-the-art method. The model we propose is also generic - it can be used to address other video understanding tasks such as action localization and audio source localization.

%
%
\bibliographystyle{splncs04}
\bibliography{main}

\clearpage

\section*{Appendix}
\label{sec:supp}
\appendix
\section{Network architecture details}
For clarification, network architecture of SPELL is shown in Table~\ref{tab:arch}.
\begin{table}[htbp]
\centering
\caption{Detailed architecture of SPELL. Batch normalization~\cite{ioffe2015batch} and ReLU~\cite{nair2010rectified} follow after each of (6), (7), and (9-14).}
\resizebox{0.85\linewidth}{!}{
\begin{tabular}{C{1.2cm}|C{2.8cm}|C{5.6cm}|C{2cm}}
\toprule
\textbf{Index} & \textbf{Inputs} & \textbf{Description} & \textbf{Dimension} \\
\midrule
(1) & - & 4-D spatial feature & 4 \\
(2) & - & Visual feature $v_{{\sf visual}}$ & 512 \\
(3) & - & Audio feature $v_{{\sf audio}}$ & 512 \\
(4) & (1) & Linear (4 $\rightarrow$ 64) & 64 \\
(5) & (2), (4) & Concatenation & 576 \\
(6) & (5) & Linear (576 $\rightarrow$ 64) & 64 \\
(7) & (3) & Linear (512 $\rightarrow$ 64) & 64 \\
(8) & (6), (7) & Addition & 64 \\
(9) & (8) & $\sf EDGE\text{-}CONV$ (Forward) & 64 \\
(10) & (8) & $\sf EDGE\text{-}CONV$ (Undirected) & 64 \\
(11) & (8) & $\sf EDGE\text{-}CONV$ (Backward) & 64 \\
(12) & (9) & $\sf SAGE\text{-}CONV$ (Forward, Shared) & 64 \\
(13) & (10) & $\sf SAGE\text{-}CONV$ (Undirected, Shared) & 64 \\
(14) & (11) & $\sf SAGE\text{-}CONV$ (Backward, Shared) & 64 \\
(15) & (12) & $\sf SAGE\text{-}CONV$ (Forward) & 1 \\
(16) & (13) & $\sf SAGE\text{-}CONV$ (Undirected) & 1 \\
(17) & (14) & $\sf SAGE\text{-}CONV$ (Backward) & 1 \\
(18) & (15), (16), (17) & Addition & 1 \\
(19) & (18) & Sigmoid & 1 \\
\bottomrule
\end{tabular}
\label{tab:arch}
}
\end{table}

\section{Qualitative results}
\begin{figure}[htbp]
  \small
    \includegraphics[width=\linewidth]{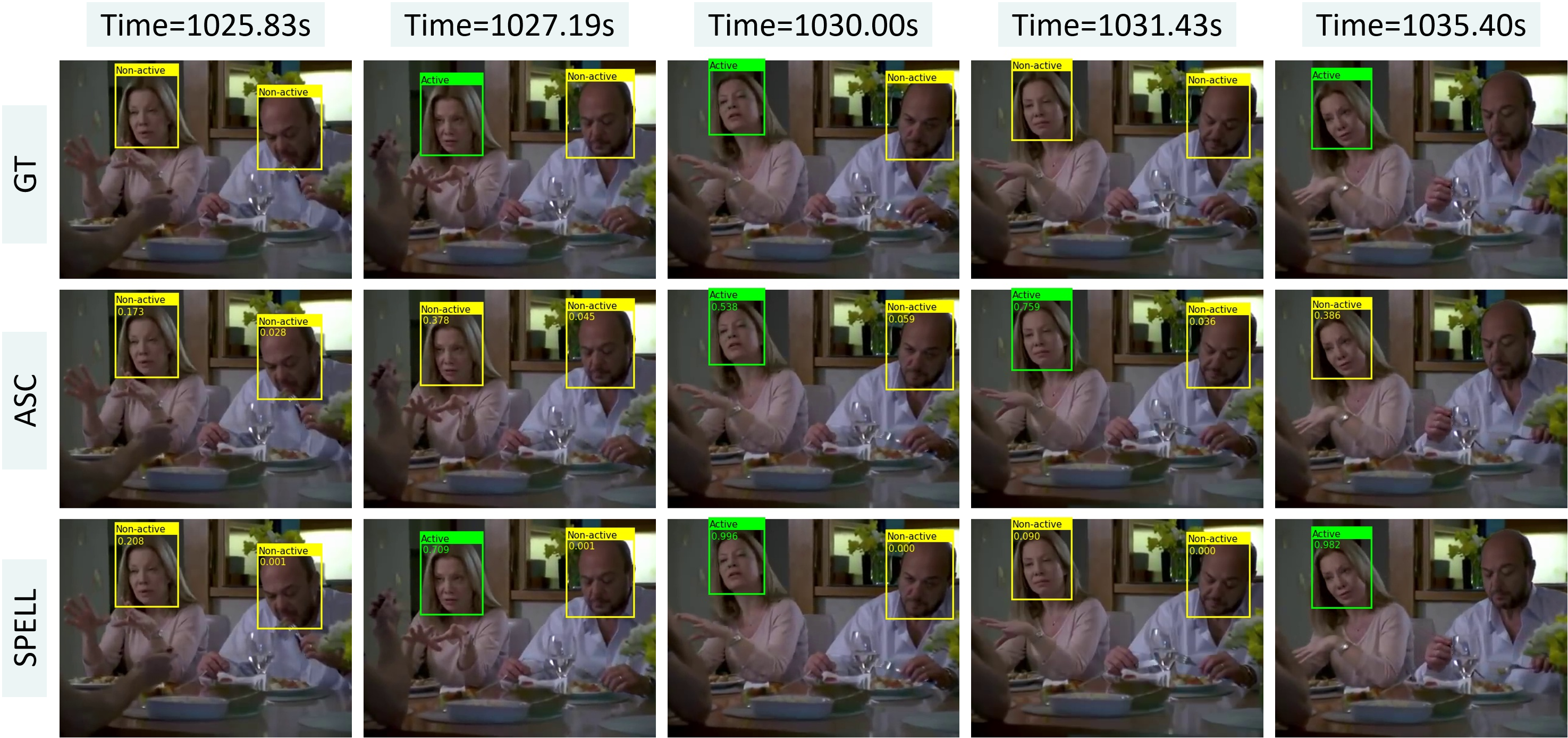} \\[2ex]
    \includegraphics[width=\linewidth]{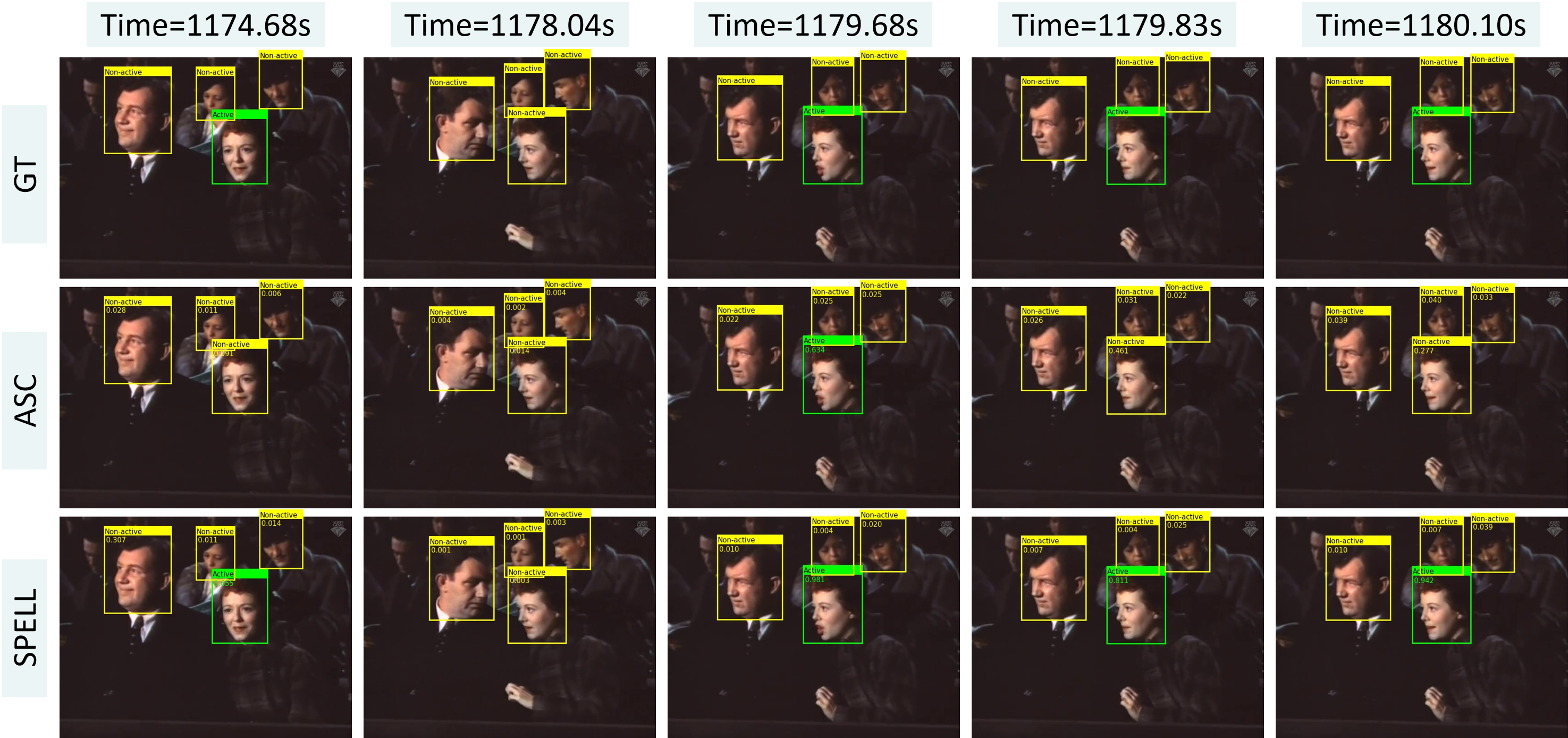} \\[2ex]
    \includegraphics[width=\linewidth]{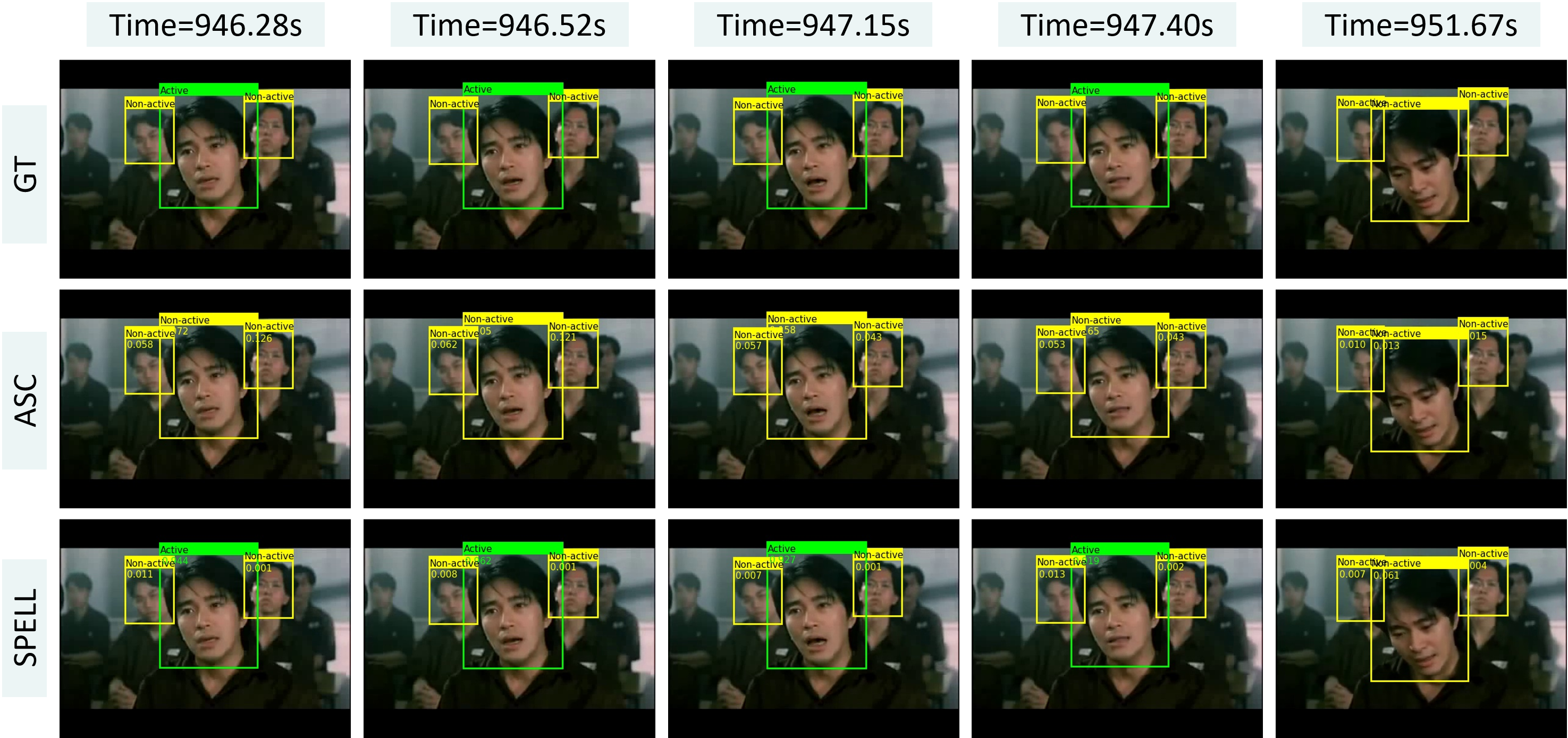}
    \caption{Qualitative results: The green boxes indicate the active speakers and yellow indicates non-active speaker. The selected frames have a long time-span about 5-10 seconds. GT: Ground Truth.}
  \label{fig:quality}
\end{figure}

In Figure~\ref{fig:quality}, we provide qualitative results of SPELL. For additional comparison, we also show the results of ASC~\cite{alcazarActiveSpeakersContext2020}, which performs the best among the approaches that have released their model weights as well as the source code. In the first example, ASC has both false positive and false negative results while SPELL is able to correctly classify all the speakers. In the second and the third examples, SPELL finds every active speaker when ASC has many false negatives. The results show that SPELL is effective and is good at modeling spatial-temporal long-term information. For more examples, please refer to the videos that are included in the subfolder.

\end{document}